\documentclass{amsart}
\usepackage{hyperref}
\usepackage{graphicx}

\begin{document}

\title{Evolving winning strategies for Nim-like games}

\author{Mihai Oltean}
\address{Department of Computer Science,\\
Faculty of Mathematics and Computer Science,\\
Babes-Bolyai University, Kogalniceanu 1,\\
Cluj-Napoca, 3400, Romania.}
\email{mihai.oltean@gmail.com}


\begin{abstract}

An evolutionary approach for computing the winning 
strategy for Nim-like games is proposed in this paper. The winning strategy 
is computed by using the Multi Expression Programming (MEP) technique - a 
fast and efficient variant of the Genetic Programming (GP). Each play 
strategy is represented by a mathematical expression that contains 
mathematical operators (such as +, -, *, mod, div, and , or, xor, not) and 
operands (encoding the current game state). Several numerical experiments 
for computing the winning strategy for the Nim game are performed. The 
computational effort needed for evolving a winning strategy is reported. The 
results show that the proposed evolutionary approach is very suitable for 
computing the winning strategy for Nim-like games.

\end{abstract}

\maketitle

\section{Introduction}

\textbf{\textit{Nim}} is one of the older two-person games known today. 
While the standard approaches for determining winning strategies for \textit{Nim} are 
based on the Grundy-Sprague theory (Berlekamp 1982, Conway 1976), 
this problem can be solved using other techniques. For instance, the first 
winning strategy for this game was proposed in 1901 by L.C. Bouton from the 
Harvard University. The Bouton's solution is based on computing the \textit{xor} sum of 
the numbers of objects in each heap. In other words Bouton computed a 
relation between the current state of the game and the player which has a 
winning strategy if it is his/her turn to move.

In this paper, we propose an evolutionary approach for computing the winning 
strategy for \textit{Nim}-like games. The proposed approach is based on Multi Expression 
Programming \footnote{MEP 
source code is available from \url{https://mepx.org} or \url{https://mepx.github.io}.} (Oltean 2003a, Oltean 2003b), which is a fast and efficient alternative to Genetic Programming (GP) (Koza 1992). The idea is to find a mathematical relation (an expression) 
between the current game state and the winner of the game (assuming that 
both players do not make wrong moves). The searched expression should 
contain some mathematical operators (such as +, -, *, \textbf{\textit{div}}, 
\textbf{\textit{mod}}, \textbf{\textit{and}}, \textbf{\textit{or}}, 
\textbf{\textit{not}}, \textbf{\textit{xor}}) and some operands (encoding 
the current game state).

It is widely known (Berlekamp 1982, Gardner 1988) that a winning 
strategy is based on separation of the game's states in two types of 
positions: $P$-positions (advantage to the previous player) and $N$-positions 
(advantage to the next player). Our aim is to find a formula that is able to 
detect whether a given game position belongs to $P$-positions or to 
$N$-positions. Our formula has to return 0 if the given position is a 
$P$-position and a nonzero value otherwise. That could be easily assimilated to 
a symbolic regression (Koza 1992) or a classification task. It is 
well-known that machine learning techniques (such as Neural Networks or 
Evolutionary Algorithms (Goldberg 1989) are very suitable for solving this 
kind of problems. However, the proposed approach is different from the 
classical approaches mainly because the $P$ and $N$-positions are usually 
difficult to be identified for a new game. Instead we propose an approach 
that checks $P$ and $N$-position during the traversing of the game tree.

This theory can be easily extended for other games that share several 
properties with the \textit{Nim} game (i.e. games for which the winning strategy is 
based on $P$ and $N$-positions).

The problem of finding $N$ and $P$-positions could be also viewed as a 
classification task with two classes. However, we do not use this approach 
because in this case is required to know the class ($P$ or $N)$ for each game 
position.

The paper is organized as follows. The \textit{Nim} game is briefly described in section 
2. MEP technique is briefly described in section 3. The fitness assignment 
process is in detail described in section 3.4. Several numerical experiments 
are performed in section 4. Section 5 outlines the possibility of using the 
proposed technique for discovering winning strategies for other games.

\section{Basics on Nim Game}

\textit{Nim} is one of the oldest and most engaging of all two-person mathematical games 
known today (Berlekamp 1982, Conway 1976). The name and the 
complete theory of the game were invented by the professor Charles Leonard 
Bouton from Harvard University about 100 years ago.

Players take turns removing objects (counters, pebbles, coins, pieces of 
paper) from heaps (piles, rows, boxes), but only from one heap at a time. In 
the normal convention the player who removes the last object wins.

The usual practice in impartial games is to call a hot position 
($N$-position - advantage to the next player, i.e. the one who is about to make 
a move) and a cold one ($P$-position - advantage to the previous player, i.e. 
the one who has just made a move).

In 1930, R. P. Sprague and P. M. Grundy developed a theory of impartial 
games in which \textbf{\textit{Nim}} played a most important role. According 
to the Sprague-Grundy theory every position in an impartial game can be 
assigned a Grundy number which makes it equivalent to a 
\textbf{\textit{Nim}} heap of that size. The Grundy number of a position is 
variously known as its \textit{Nim-heap} or \textit{nimber} for short (Berlekamp 1982, Conway 1976).

A P-position for the \textbf{\textit{Nim}} game is given by the equation:

$x_{1}$ \textbf{\textit{xor}} $x_{2}$ \textbf{\textit{xor}} \ldots 
\textbf{\textit{xor}} $x_{n}$ = 0,

\noindent
where $n$ is the number of heaps, $x_{i}$ is the number of objects in the 
$i^{th}$ heap and \textbf{\textit{xor}} acts as the \textbf{modulo} 2 
operator.

\section{Multi Expression Programming Technique}

The MEP representation, evolutionary scheme and the fitness assignment 
process are briefly described in this section.

\subsection{Individual Representation}

MEP representation is similar to the way in which \textbf{\textit{C}} or 
\textbf{\textit{Pascal}} compilers translate mathematical into machine code 
(Oltean 2003). 
\textbf{\textit{Pascal}} or \textbf{\textit{C}} compilers use so called 
three addresses code (or intermediary code) when they translate an infix 
form mathematical expression (i.e. the human readable form) into an 
executable machine code.

MEP genes are substrings of variable length. Number of genes in a chromosome 
is constant and it represents \textit{chromosome length}. Each gene encodes a terminal or a function 
symbol. A gene encoding a function includes pointers towards the function 
arguments. Function parameters always have indices of lower values than the 
position of that function itself in the chromosome. 

According to the proposed representation scheme the first symbol of the 
chromosome must be a terminal symbol.\\

\textbf{Example}\\

An example of chromosome $C_{MEP}$ is given below:\\

A representation where numbers on the left positions stand for gene labels 
(or memory addresses) it is used. Labels do not belong to the chromosome. 
They are provided for explanation purposes only. \\

1: $a $

2: $b$

3: + 1, 2

4: $c$

5: $d$

6: * 3, 5\\

When MEP individuals are translated into (expressions) computer programs 
they are read downstream starting with the first position. A terminal symbol 
specifies a simple expression. A function symbol specifies a complex 
expression (formed by linking the operands specified by the argument 
positions with the current function symbol).

For instance, genes 1, 2, 4 and 5 in previous example encode simple 
expressions composed of a single terminal symbol. The expressions associated 
to the genes 1, 2, 4 and 5 are:\\

$E_{1}=a$, 

$E_{2}=b$, 

$E_{4}=c$, 

$E_{5}=d$, \\

Gene 3 indicates the operation + on the operands located at positions 1 and 
2 of the chromosome. Therefore gene 3 encodes the expression:\\

$E_{3}=a+b$.\\

Gene 6 indicates the operation * on the operands located at positions 3 and 
5. Therefore gene 6 encodes the expression:\\

$E_{6}$ = ($a + b)$ * $d$.\\

The expression encoded into a MEP chromosome is the expression encoded by 
its last gene. Thus, the expression encoded by the previously described MEP 
chromosome is ($a + b)$ * $d$.

\subsection{MEP for Evolving Nim Strategies}

In order to use MEP technique for evolving a formula that may be used for 
identifying $P$ and $N$-positions we have to define the terminal symbols and the 
function symbols. Note that these sets strongly depend on the problem being 
solved. For instance if we tackle the \textit{Nim} game, the set $T$ may consists of the 
following values: number of heaps ($n)$ and the number of objects in each heap. 
Thus $T$ = {\{}$n$, $a_{1}$, $a_{2}$, \ldots , $a_{n}${\}}. The set $F$ is more general 
and it usually contains some mathematical operators. In this paper we use 
the set $F$ = {\{}+, -, *, \textbf{\textit{div}}, \textbf{\textit{mod}}, 
\textbf{\textit{and}}, \textbf{\textit{or}}, \textbf{\textit{xor}}, 
\textbf{\textit{not}}{\}}. \\

\textbf{\textit{Remarks}}: \\

\noindent
i) For the \textbf{\textit{Nim}} game only the operator 
\textbf{\textit{xor}} is needed. However, to avoid biases an extended set of 
operators has been used.

\noindent
ii) \textbf{\textit{and}}, \textbf{\textit{or}}, \textbf{\textit{xor}} and 
\textbf{\textit{not}} are interpreted as bitwise operators.

\subsection{Genetic Operators}

Genetic operators that may be used in conjunction with the MEP 
technique are one-point crossover and mutation.\\

\textbf{One-point crossover}\\

By applying recombination operator, one crossover point is randomly chosen 
and the parents exchange the sequences after the crossover point.\\

\textbf{Mutation}\\

Each MEP gene may be subject of mutation. To preserve the consistency of the 
chromosome its first gene must encode a terminal symbol. For other genes 
there is no restriction in symbols changing.

If the current gene encodes a terminal symbol it may be changed into another 
terminal symbol or into a function symbol. In the last case the positions 
indicating the function arguments are also generated by mutation.

If the current gene encodes a function the gene may be mutated into a 
terminal symbol or into another function (function symbol and pointers 
towards arguments).

\subsection{Fitness Assignment Process}

In this section the procedure used for computing the quality of a chromosome 
is described.

Even if this problem could be easily handled as a classification problem 
(based on a set of fitness cases), we do not use this approach since for the 
new games it is difficult to find which the $P$-positions and $N$-positions are. 
Instead we employ an approach based on the traversing the game tree. Each 
nod in this tree is a game configuration (state).

There are three theorems that run the winning strategy for the 
\textbf{\textit{Nim}} game (Berlekamp 1988):

\begin{itemize}

\item[{\it (i)}]{any move applied to a $P$-position turns the game into a $N$-position,}

\item[{\it (ii)}]{there is at least one move that turns the game from a $N$-position into a 
$P$-position,}

\item[{\it (iii)}]{the final position (when the game is over) is a $P$-position.}

\end{itemize}

The value of the expression encoded into a MEP chromosome is computed for 
each game state. If the obtained value is 0, the corresponding game state is 
considered as being a $P$-position, otherwise the configuration is considered 
as being a $N$-position.

The fitness of a chromosome is equal to the number of violations of the 
above described rule that arises in a game tree. Thus, if the current 
formula (chromosome) indicates that the game state encoded into a node of 
the game tree is a $P$-position and (the same current formula indicates that) 
all the game states encoded in the offspring nodes are also $P$-positions means 
that we have a violation of the rule $b)$.

Since we do not want to have violations of the previously described rule, 
our chromosome must have the fitness equal to zero. This means that the 
fitness has to be minimized.

For a better understanding of the fitness assignment process we provide an 
example where we shall compute by hand the fitness of a chromosome.

\begin{figure}[htbp]
\centerline{\includegraphics[width=2.52in,height=1.95in]{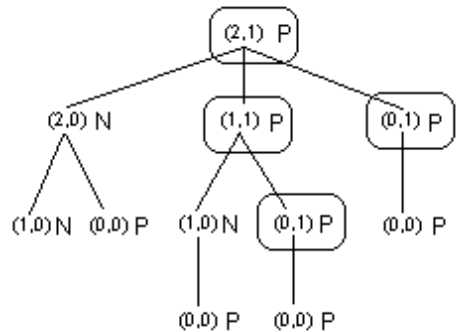}}
\label{nim_fig1}
\caption{The game tree for a Nim game that starts with the 
configuration (2, 1). At the right side of each game configuration is 
printed the configuration' state ($P$-position or $N$-position) as computed by the 
formula $E=a_{1} - a_{1}$*$a_{2}$. The configurations that violate one of 
the three rules described above are encircled.}
\end{figure}

Consider the game state (2,1), and a MEP chromosome encoding the expression 
$E=a_{1} - a_{2}$*$a_{1}$. The game tree of the \textbf{\textit{Nim}} game 
is given in Figure \ref{nim_fig1}.

Figure \ref{nim_fig1} shows that the fitness of a MEP chromosome encoding the formula $E$ = 
$a_{1} - a_{2}$*$a_{1}$ is four (there are four violations of the winning 
strategy rules).

\subsection{The Evolutionary Algorithm}

MEP uses a steady-state (Syswerda 1989) evolutionary scheme. Initial 
population is randomly generated. The following steps are repeated until a 
termination criterion is reached: Two parents are selected (from 4 
individuals) using binary tournament and are recombined in order to obtain 
two offspring. The offspring are considered for mutation. The best offspring 
replaces the worst individual in the current population if the offspring is 
better than the worst individual.

\section{Numerical Experiments}

Several numerical for evolving winning strategies for \textit{Nim}-like games are 
performed in this section.

The purpose of these experiments is to evolve a formula capable to 
distinguish between a $N$-position and a $P$-position for the 
\textbf{\textit{Nim}} game. We shall analyze the relationships between the 
success rate and the population size, the chromosome length and the number 
of generations used during the search process. 

In all the experiments it is considered the following configuration for the 
\textbf{\textit{Nim}} game: (4, 4, 4, 4). This configuration has been chosen 
in order to have a small computational time. However, this configuration has 
proved to be enough for evolving a winning strategy.

The total number of game configurations is 70 (which can be obtained either 
by counting nodes in the game tree or by using the formula of combinations 
with repetitions). Two permutations of the same configuration are not 
considered different.

General parameter setting for the MEP algorithm are given in Table \ref{nim_tab1}.

\begin{table}[htbp]
\label{nim_tab1}
\caption{General paramters for the MEP algorithm.}
\begin{center}
\begin{tabular}
{p{140pt}p{150pt}}
\hline
\textbf{Parameter}& 
\textbf{Value} \\
\hline
Chromosome length& 
15 genes \\
Number of generations& 
100 \\
Crossover probability& 
0.9 \\
Mutations& 
2 mutations / chromosome \\
Selection strategy& 
binary tournament \\
Terminal set& 
$T_{Nim}$ = {\{}$n$, $a_{1}$, $a_{2}$, \ldots , $a_{n}${\}}. \\
Function set& 
$F $= {\{}+, -, *, \textbf{\textit{div}}, \textbf{\textit{mod}}, \textbf{\textit{and}}, \textbf{\textit{not}}, \textbf{\textit{xor}}, \textbf{\textit{or}}{\}} \\
\hline
\end{tabular}
\end{center}
\end{table}

\textbf{\textit{Remark}}: The success rate is computed by using the formula:

\[
\mbox{Success}\,\mbox{rate} = \frac{\mbox{the number of successful 
}\,\mbox{runs}}{\mbox{the total number of runs}}.
\]

\textbf{Experiment 1}\\

In the first experiment the relationship between the population size and the 
success rate is analyzed. Each MEP chromosome has 15 genes and the search process was evolved for 100 generations. Other MEP parameters are given in Table \ref{nim_tab1}.

\begin{figure}[htbp]
\centerline{\includegraphics[width=3.02in,height=2.49in]{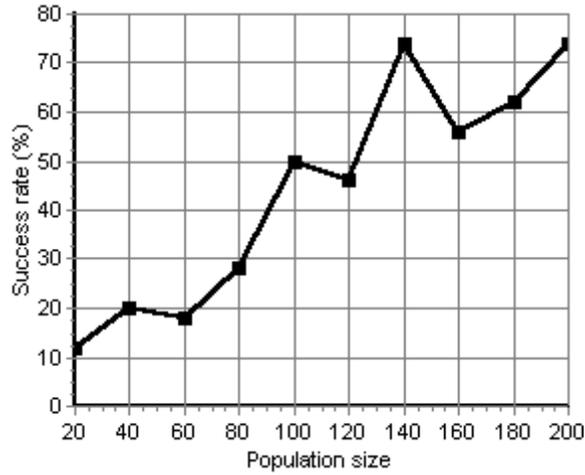}}
\label{nim_fig2}
\caption{The relationship between the population size and the rate 
of success. The results are averaged over 50 runs. The population size 
varies between 20 and 200.}
\end{figure}

The results of this experiment are depicted in Figure \ref{nim_fig2}.

Figure \ref{nim_fig2} shows that the success rate increases as the 
population size increases. The highest value - 37 successful runs (out of 
50) - is obtained with a population containing 140 individuals. Even a 
population with 20 individuals is able to yield 6 successful runs (out of 
50).\\

\textbf{Experiment 2}\\

In the second experiment the relationship between the number of generations 
and the success rate is analyzed. A population with 100 individuals each having 15 genes is used in this experiment. Other MEP parameters are given in Table \ref{nim_tab1}.

\begin{figure}[htbp]
\centerline{\includegraphics[width=3.01in,height=2.50in]{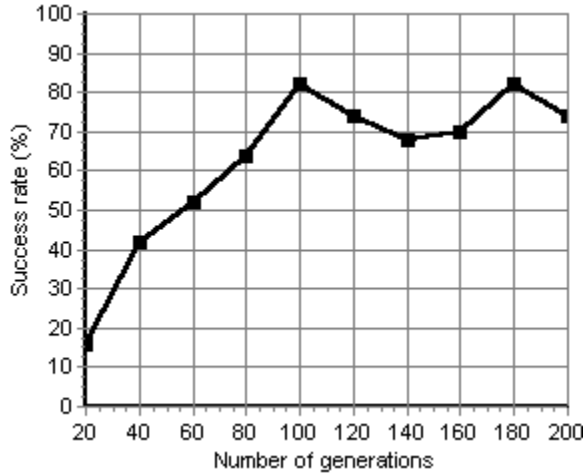}}
\label{nim_fig3}
\caption{The relationship between the number of generations and 
the rate of success. The results are averaged over 50 runs. The number of 
generations varies between 20 and 200.}
\end{figure}

The results of this experiment are depicted in Figure \ref{nim_fig3}.

From Figure \ref{nim_fig3} it can be seen that MEP is able to find a winning strategy for 
the \textbf{\textit{Nim}} game in most of the runs. In 41 runs (out of 50) a 
perfect solutions was obtained after 100 generations. 9 successful runs were 
obtained when the algorithm is run for 20 generations.\\

\textbf{Experiment 3}\\

\begin{figure}[htbp]
\centerline{\includegraphics[width=2.97in,height=2.46in]{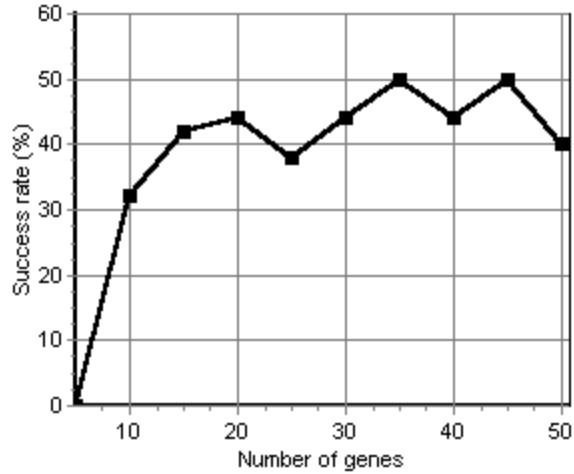}}
\label{nim_fig4}
\caption{The relationship between the chromosome length and the 
success rate. The results are averaged over 50 runs. The chromosome length 
varies between 5 and 50.}
\end{figure}

In the third experiment the relationship between the chromosome length and 
the success rate is analyzed. A population with 100 individuals is evolved for 50 generations. Other MEP parameters are given in Table \ref{nim_tab1}. 
The results of this experiment are depicted in Figure \ref{nim_fig4}.

Figure \ref{nim_fig4} shows that the optimal number of genes of a MEP 
chromosome is 35. With this value 25 runs (out of 50) were successful. It is interesting to note that the formulas evolved by MEP are sometimes 
different from the classical $a_{1}$ \textbf{\textit{xor}} $a_{2}$ 
\textbf{\textit{xor}} $a_{3}$ \textbf{\textit{xor}} $a_{4}$. For instance a 
correct formula evolved by MEP is:

$F=a_{1}$ \textbf{\textit{xor}} $a_{2}$ \textbf{\textit{xor}} $a_{3} - a_{4}$.

This formula is also correct due to the properties of the 
\textbf{\textit{xor}} operator.

\section{Conclusions}

In this paper, an evolutionary approach for the \textbf{\textit{Nim}} game 
has been proposed. The underlying evolutionary technique is \textit{Multi Expression Programming} - a very fast 
and efficient \textit{Genetic Programming} variant. Numerical experiments have shown that MEP is able to discover a winning strategy in most of the runs.

The proposed method can be easily applied for games whose winning strategy 
is based on $P$ and $N$-positions. The idea is to read the game tree and to count 
the number of configurations that violates the rules of the winning 
strategy.

\end{document}